\pdfoutput=1

\documentclass[11pt]{article}

\usepackage[]{ACL2023}

\usepackage{times}
\usepackage{latexsym}

\usepackage[T1]{fontenc}

\usepackage[utf8]{inputenc}

\usepackage{microtype}

\usepackage{inconsolata}

\usepackage{microtype}
\usepackage{graphicx}
\usepackage{bm}
\usepackage{amsmath}
\usepackage{amsfonts}
\usepackage{bbm}

\usepackage{helvet}
\usepackage{courier}
\usepackage{multirow}
\usepackage{bm}
\usepackage{amsmath}
\usepackage{mdwlist}

\usepackage{relsize}
\usepackage{amsfonts}
\usepackage{url}
\usepackage{graphicx}
\usepackage{subfigure}
\usepackage{caption}
\usepackage{hhline}
\usepackage{xcolor}
\usepackage{color}
\usepackage{colortbl}
\usepackage{booktabs}
\usepackage{tabularx}
\usepackage{amsmath}
\usepackage[linesnumbered,boxed,ruled,commentsnumbered]{algorithm2e}

\title{Actively Supervised Clustering for Open Relation Extraction}

\author{Jun Zhao$^{1}$\footnotemark[1],\ \ Yongxin Zhang$^{1}$\footnotemark[1],\ \ Qi Zhang$^{1}$\footnotemark[2],\ \ Tao Gui$^{2}$\footnotemark[2], \\ \textbf{Zhongyu Wei}$^3$\textbf{,}\ \ \textbf{Minlong Peng}$^4$\textbf{,}\ \ \textbf{Mingming Sun}$^4$\\
  $^1$School of Computer Science, Fudan University\\
  $^2$Institute of Modern Languages and Linguistics, Fudan University\\
  $^3$School of Data Science, Fudan University\\
  $^4$Cognitive Computing Lab Baidu Research\\
  \texttt{\{zhaoj19,yongxinzhang20,qz,tgui,zywei\}@fudan.edu.cn}}

\begin{document}
\maketitle
\renewcommand{\thefootnote}{\fnsymbol{footnote}}
\footnotetext[1]{Equal Contributions.}
\footnotetext[2]{Corresponding authors.}
\begin{abstract}

Current clustering-based Open Relation Extraction (OpenRE) methods usually adopt a two-stage pipeline. The first stage simultaneously learns relation representations and assignments. The second stage manually labels several instances and thus names the relation for each cluster.
However, unsupervised objectives struggle to optimize the model to derive accurate clustering assignments, and the number of clusters has to be supplied in advance.
In this paper, we present a novel setting, named actively supervised clustering for OpenRE. Our insight lies in that clustering learning and relation labeling can be alternately performed, providing the necessary guidance for clustering without a significant increase in human effort. 
The key to the setting is selecting which instances to label. Instead of using classical active labeling strategies designed for fixed known classes, we propose a new strategy, which is applicable to dynamically discover clusters of unknown relations. Experimental results show that our method is able to discover almost all relational clusters in the data and improve the SOTA methods by 10.3\% and 5.2\%, on two datasets respectively.

\end{abstract}

\section{Introduction}

\noindent Relation extraction (RE) aims to detect and extract the potential relation between the given entity pair in unstructured text. The extracted relation facts play a vital role in many downstream applications, such as knowledge base population \cite{ji2011knowledge}, search engine \cite{10.1007/978-3-319-93417-4_38}, and question answering \cite{yu2017improved}. 
To deal with the emerging unknown relational types in the real world, Open Relation Extraction (OpenRE) has been widely studied.

    \begin{figure}[t]
        \includegraphics[width=\columnwidth]{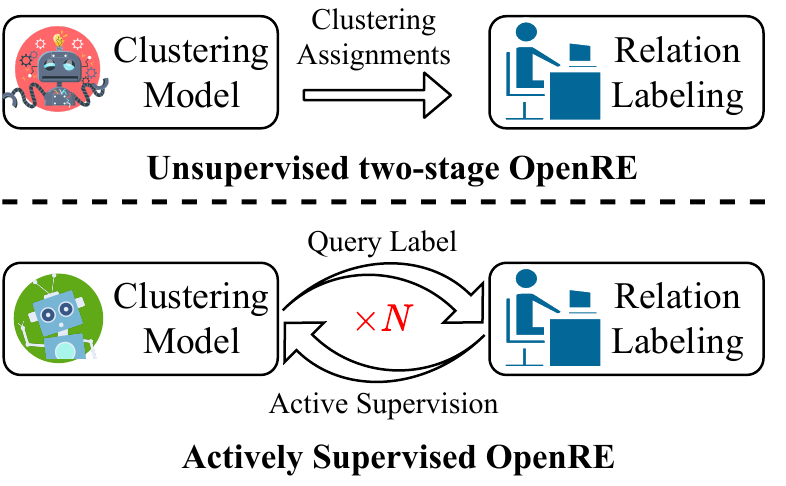}
        \caption{Compared with the existing unsupervised two-stage methods, our method can provide explicit supervision for clustering by alternately performing clustering learning and relation labeling. Note that the human effort of the two settings is comparable.}
        \label{fig:intro}
    \end{figure}

The clustering-based unsupervised relation discovery is a classical paradigm for OpenRE  \cite{yao-etal-2011-structured,10.1162/tacl_a_00095,elsahar2017unsupervised}. It can discover potential relations, by grouping several instances into relational clusters, and then manually labeling a few instances to name the relation of each cluster. 
Recently, \citet{hu-etal-2020-selfore} introduced a deep clustering framework \cite{caron2018deep} into OpenRE. They iteratively cluster the relation representations that are produced by large pretrained models and use the cluster assignments as pseudo-labels to refine the representations. Unfortunately, the above unsupervised methods struggle to learn good enough representations, and the cluster assignments are error-prone. When multiple relations are mixed in a cluster, it becomes difficult to name the cluster. Hence, instead of regarding OpenRE as a totally unsupervised task, researchers leverage the labeled data of predefined relations to provide explicit supervision signals for clustering learning \cite{wu2019open,zhao2021rocore}, and achieve superior results.

Different from the above two-stage methods, in this work, we present a new setting named actively supervised clustering for OpenRE (ASCORE). As shown in fig. \ref{fig:intro}, our insight lies in that clustering learning (i.e, deep clustering) and relation labeling can be alternately performed. In an iteration, a small number of key instances are selected for labeling. The unknown relations expressed by these instances are correspondingly discovered. More importantly, these labeled instances can provide explicit supervisory signals for clustering learning. The improved relation representations form a better cluster structure, which in turn is able to benefit the discovery of the neglected relations. Since potential relations are dynamically discovered in iterations, the number of clusters does not need to be provided in advance. 

Along with this setting, we design an active labeling strategy tailored for clustering. First, all instances are encoded to points in the representation space, where the clustering is performed. The goal of the strategy is to select the most informative points for labeling. Intuitively, two points that are far from each other in representation space usually express different relations. To discover as many relations as possible, we introduce a distance regularization to the strategy, so that diversified relation discovery can be facilitated. To prevent over-fitting caused by training with limited active labeled instances, all the selected key points are required to be the points of maximum local density. By doing so, a large number of high-quality pseudo-labels can be obtained, by assigning active labels to unlabeled data in a small neighborhood. To mitigate the error propagation issue, different loss functions are assigned to active labels and pseudo-labels with different reliability for clustering learning.
Experimental results show that (1) the actively supervised method improves the SOTA two-stage methods by a large margin without a significant increase in human effort. (2) the proposed active strategy can discover more relational clusters, compared with the classical active strategy.

To summarize, the main contributions of this work are as follows: (1) We present a new setting named actively supervised clustering for OpenRE, providing the necessary guidance for clustering without a significant increase in human effort.
(2) Design of a new active labeling strategy tailored for clustering, that can effectively discover potential relational clusters in unlabeled data.
(3) This method improves the SOTA two-stage methods by 10.3\% and 5.2\% on two well-known datasets, respectively.

\section{Related Work}
\noindent\textbf{Clustering-based OpenRE}: 
The clustering-based paradigm considers relation discovery as a two-stage pipeline, which clusters relational data first, and then manually labels relational semantics for each cluster. Conventional methods cluster instances by human-defined linguistic features \cite{yao-etal-2011-structured,10.1162/tacl_a_00095,elsahar2017unsupervised}, such as entity words/type, dependency paths, trigger words, context POS tags. Recently, many studies have shown that pretrained models learn diversified linguistic knowledge \cite{jawahar-etal-2019-bert,clark-etal-2019-bert,Goldberg2019AssessingBS,zhao-etal-2022-read-extensively}. \citet{hu-etal-2020-selfore} leverage the self-supervised signals provided by the pretrained model to iteratively learn relation representations and optimize clustering. Due to the lack of strong supervision, it is difficult for the above methods to produce satisfactory clustering results. Although some works \cite{wu2019open,zhao2021rocore} try to use the labeled data of predefined relations to complete the missing supervision, the semantic gap between predefined and open relations leads to negative clustering bias, especially when these relations come from different domains \cite{zhao2021rocore}. By performing clustering learning and relation labeling alternately, our actively supervised method can provide strong supervision and improve the two-stage methods by a large margin. \textit{In the main results (sec. \ref{sec:main_res}), we achieve this improvement at the cost of only two active labels for each relation on average. For two-stage methods, relation labeling for each cluster requires at least one (usually more) instance to be manually observed.} Therefore, there is no significant increase in human effort.

    \begin{figure*}[t]
    \centering
        \includegraphics[width=\linewidth]{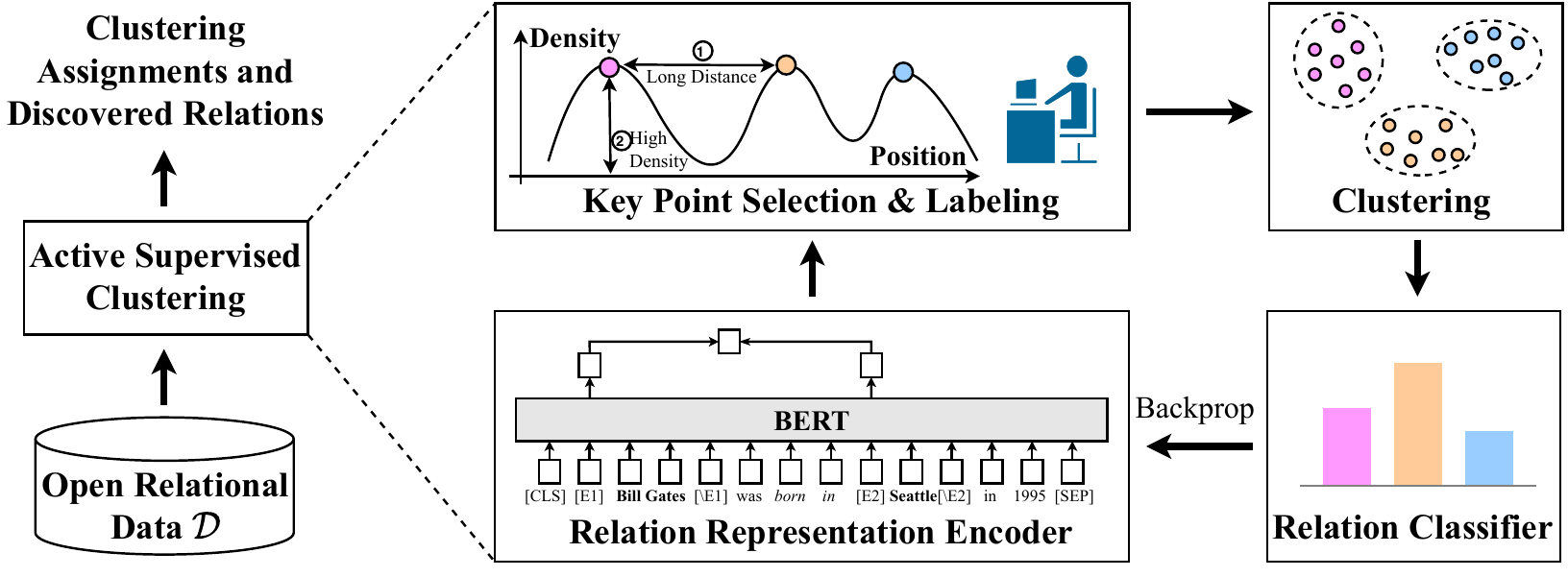}
        \caption{Overview of the training pipeline for our actively supervised clustering setting. In each iteration, a few key points are selected for relation labeling. The rest instances are clustered to the nearest key points. Some highly reliable cluster assignments are used as pseudo-labels for relation representation learning.}
        \label{fig:model}
    \end{figure*}  
\noindent\textbf{Active Learning}:
Active learning is a research field with high relevance to the proposed methods.
In the research field, a classical method is uncertainty-based sampling\cite{10.1007/11871842_40,6889457,tong2001support}. The uncertainty can be defined based on the posterior probability of a predicted class or the distances to the decision boundaries.
In the context of deep learning, MC Dropout \cite{gal2017deep} is an effective way for uncertainty estimation, but the computationally inefficient limits its application in large-scale datasets. Recently, representative sampling is attracting lots of attention \cite{sener2018active,ash2019deep}, which selects data points that represent the distribution of an unlabeled pool. \textit{Unlike the classical labeling strategies designed for fixed classes, the proposed strategy is encouraged to discover new relational clusters while improving the clustering of relations that have been discovered.}

\section{Approach}
    In this work, we present a new setting named actively supervised clustering for OpenRE (ASCORE), which fuses the isolated two-stage pipeline to guide clustering learning. Fig. \ref{fig:model} illustrates the training pipeline. The OpenRE problem addressed in this work is formally stated as follows. Given as input an open relational dataset $\mathcal{D}=\{x_i|i=1,..,N\}$, the goal is to discover and label potential relations $\mathcal{R}=\{r_i|i=1,..,K\}$ in the open data and cluster the corresponding instances. Note that the number of relations $K$ in $\mathcal{D}$ is \textit{unknown}.

    \subsection{Overview}
        The ASCORE is based on deep clustering \cite{caron2018deep}, which is a common practice for OpenRE \cite{hu-etal-2020-selfore,zhao2021rocore}, that iteratively clusters the representations of input instances, and uses the cluster assignments as pseudo-labels to learn the relation representations. We introduce explicit supervision to deep clustering by alternately performing clustering learning and relation labeling. The actively labeled points can serve as a basis to facilitate accurate pseudo-label estimation and improve representation learning. The improved relational representation in turn benefits relation labeling to discover more relations. 
        As illustrated in Figure \ref{fig:model}, the training pipeline of ASCORE consists of the following steps:

        \textbf{Encoding Step}: This step aims to obtain the relation representation $\bm{h}_i$ of each input instance $x_i\in\mathcal{D}$, laying the groundwork for clustering and relation discover.  First, the contextual information of $x_i$ is encoded to the entity pair representation $\bm{h}_i^{ent}$, using a pretrained BERT \citep{DBLP:journals/corr/abs-1810-04805} encoder. To avoid data sparsity and low efficiency of clustering in high-dimensional space, an autoencoder is used to transform $\bm{h}_i^{ent}$ into a low-dimensional clustering-friendly representation $\bm{h}_i$.

        \textbf{Labeling Step}: This step aims to discover potential relations in open dataset $\mathcal{D}$ and guide clustering. At the $s^{\text{th}}$ iteration, a set $\mathcal{D}^*_s\in\mathcal{D}$ is actively labeled, including $B$ key points. These key points are required to be local maximum in density, so a large number of high-quality pseudo labels can be obtained by assigning active labels to unlabeled data in a small neighborhood. Instead of focusing only on the improving clustering of the discovered relations, all key points in $\mathcal{D}^*=\mathcal{D}^*_1\cup..\cup\mathcal{D}^*_s$ are required to be far away from each other to facilitate the discovery of new relations.

        \textbf{Learning Step}: This step aims to learn clustering relational data, using an actively labeled $\mathcal{D}^*$. Specifically, each unlabeled point $x_i \in \mathcal{D}$ is clustered to the nearest key point $x_j^*\in \mathcal{D}^*$ and the pseudo label is $\hat{y}_i=y_j^*$. The reliability of $\hat{y}_i$ increases as the distance in representation space between $x_i$ and $x_j^*$ decreases. Cross-entropy loss (resp. divergence-based contrastive loss) is used for pseudo labels with high (resp. moderate) reliability, to optimize relation representations and thus to improve clustering. With the help of active supervision, separated subclusters expressing the same relation approach each other, while mixed subclusters expressing different relations are separated. Existing unsupervised methods are inherently difficult to handle such errors.
        
        The above three steps are performed iteratively to gradually improve the model performance. In the following sections, we will elaborate on the model structure, labeling strategy, and training methods involved in the above three steps.
        
    \subsection{Relation Representation Encoder} 
    \label{sec:encoding}
        Given a relation instance $x_i=\{w_1, w_2, ..., w_n\}$, in which four reserved word pieces $[E1], [\backslash E1]$, $[E2], [\backslash E2]$ are used to mark the beginning and end of each entity,
        the relation representation encoder $\bm{f}$ aims to encode the contextual relational information of instance $x_i$ into a fixed-length representation $\bm{h}^{ent}_i=\bm{f}(x_i)\in\mathbb{R}^d$. The encoder $\bm{f}$ is implemented as BERT. Specifically:
        \begin{gather}
            \bm{h}_1,...,\bm{h}_n={\rm BERT}(w_1,...,w_n)\\
            \bm{h}^{ent}=\left \langle\bm{h}_{[E1]}|\bm{h}_{[E2]}\right \rangle.
        \end{gather}
        Following \citet{soares2019matching}, the fixed-length representations are obtained, by concatenating the hidden states of marker $[E1], [E2]$ (i.e., $\bm{h}_{[E1]}$ ,$\bm{h}_{[E2]}$) and $\left\langle\cdot|\cdot\right\rangle$ is the concatenation operator. 
        
        However, clustering data in high-dimensional space is time-consuming and the data sparsity leads to sub-optimal clustering results. Therefore, an autoencoder is trained by reconstruction loss $\mathcal{L}_{rec}$ and the encoder part is retained to transform high-dimensional $\bm{h}^{ent}$ to a low-dimensional clustering-friendly relation representation $\bm{h}$.
        
    \subsection{Key Point Selection Module}
    \label{sec:labeling}
        In this section, the proposed key point selection method will be explained, including the labeling strategy and the conditions for stopping annotation.
        
        \noindent\textbf{Labeling strategy.} The labeling strategy is based on the following criteria. \textbf{First}, the selected key points are the local maximum in density. Generally, labels do not drastically change within a small neighborhood, and therefore, the first criteria enable to find a lot of unlabeled data within a small neighborhood of each key point, and accurately estimate their pseudo-labels.
        To find these local maximum, it is calculated the euclidean distance between the relation representations $\{\bm{h}_i\}_{i=1,2,...,N}$ obtained in encoding step, and the distance matrix $\bm{D}\in \mathbb{R}^{N\times N}$ is constructed as follows:
        \begin{equation}
            \bm{D}_{ij}=\left\|\bm{h}_{i}-\bm{h}_j\right\|_2^2,
        \end{equation}
        where $\bm{D}_{ij}$ is the distance between two relational instance $x_i$ and $x_j$. The potential computational cost to process large-scale datasets can be solved by sampling a small subset. Based on distance matrix $\bm{D}$, a density $\rho_i$ is further defined for each relation instance $x_i$. A larger $\rho_i$ indicates a larger number of instances around $x_i$:
        \begin{equation}
            \rho_i=\sum_{j=1}^{N}sign(D_c-D_{ij}),
        \end{equation}
        where $sign()$ is the sign function and $D_c$ is a threshold. 

        To avoid the problem that all the labeled points are concentrated in several high-density areas and missing most long-tail relations, the \textbf{second} criteria is to keep these key points away from each other in clustering space. Specifically, a sparsity index $\xi_i$ is defined for each instance $x_i\in \mathcal{D}$.
        \begin{equation}
            \xi_i=\left\{
            \begin{array}{lc}
            \min_{j,\rho_j> \rho_i}D_{ij},& {\rho_i<\rho_{max}}\\
            \max_jD_{ij}& {\rho_i=\rho_{max}}\\
            \end{array} \right.
        \end{equation}
        Intuitively, a larger $\xi_i$ indicates that the instance $x_i$ is a local maximum of density in a larger radius. 
        Based on the density $\rho_i$ and sparsity index $\xi_i$ of each instance, the labeling strategy can be formally stated as follows. \textit{In each iteration, choose $B$ points with the highest density and their distance from each other is greater than} $\xi_c$.
        \begin{gather}
            \mathcal{D}^*_s=\text{TopB}_{\rho}\{x_i|\xi_i>\xi_c,x_i\in \mathcal{D}\}
        \end{gather}

        To effectively support iterative labeling and maintain the diversity of key points, in $s^{\text{th}}$ iteration, each new key point $x_i$ should be as far away from the existing key point in $\mathcal{D}^*=\mathcal{D}^*_1\cup...\cup\mathcal{D}_{s-1}$ as possible. Therefore, for each instance $x_i$, the sparsity index are modified as follows:
        \begin{gather}
            d=\min_{x_j\in\mathcal{D}^*}||\bm{h}_i-\bm{h}_j||_2^2\\
            \xi_i=\min(\xi_i,d),
        \end{gather}
        After the $s^{\text{th}}$ iterations, the result is the new active labeled set $\mathcal{D}^*=\mathcal{D}^*\cup \mathcal{D}^*_s$.
        
    \noindent\textbf{Conditions for stopping annotation.}
        Too few queries will lead to missing some relations, while too many queries will lead to unnecessary costs. Here we give a simple strategy to determine when to stop labeling. (1) First, users can determine the maximum number of actively labeled instances, $N^*$, based on their annotation budget. (2) Since our labeling strategy takes into account the diversity of key points, new relations are constantly discovered during the initial phase of iteration. When no new relations are discovered in two or more consecutive iteration steps (it means that most relations have been found), the labeling should be stopped.

    \subsection{Training Methods}
    \label{sec:learning}
        
        \noindent\textbf{Pseudo Label Estimation.}
        Given the active labeled set $\mathcal{D}^*$, each of the rest unlabeled points $x_i\in \mathcal{D}$ is clustered to the nearest key point $x_j^*\in \mathcal{D}^*$ and $\hat{y}_i$ is estimated as $y_j^*$. Intuitively, the accuracy of the pseudo label decreases with the increase of the distance between $x_i$ and $x_j^*$. The reliability $r$ of pseudo labels is defined as follows:
            \begin{gather}
                 r_i=\left\|\bm{h}_{i}-\bm{h}_j^*\right\|_2^{-1},
                \label{equ:estimation}
            \end{gather}    
        
        where $\bm{h}_i$ and $\bm{h}_j^*$ denote the representation of $x_i$ and $x_j^*$, respectively. $\|\cdot\|_2^{-1}$ denotes reciprocal of $L_2$ norm.
        
        \noindent\textbf{Model Optimization.}
        Given the pseudo label $\hat{y}_i$ and its reliability $r_i$ for each unlabeled data $x_i \in \mathcal{D}$, the relation representation is refined to improve clustering in the next iteration. Specifically, we first filter out a highly reliable subset $\mathcal{D}_{h}=\{(x_i,\hat{y}_i)|r_i>r_{h}\}$ and use a softmax classifier to convert entity pair representation $\bm{h}^{ent}_i$ into the probability distribution on discovered relations (denoted as $\mathcal{P}_i$). The model is optimized with cross entropy loss for fast convergence:
            \begin{equation}
                \mathcal{L}_{ce}=\text{CrossEntropy}(\hat{y}_i,\mathcal{P}_i).
                \label{equ:ce}
            \end{equation}    
        Note that the number of instances in $\mathcal{D}_h$ is small. To avoid that the model only learns simple features, the threshold is broaden, and a moderately reliable subset $\mathcal{D}_m=\{(x_i,\hat{y}_i)|r_i>r_{m}\}$ containing more instances is built.
        To mitigate the negative impact of noise in $\mathcal{D}_m$, a binary contrastive loss is further introduced: 
            \begin{gather}
               \mathcal{L}_{bce}=\mathcal{L}_{\hat{y}_i=\hat{y}_j}+\mathcal{L}_{\hat{y}_i\neq\hat{y}_j}\\ \mathcal{L}_{\hat{y}_i=\hat{y}_j}=\mathcal{D}_{kl}(\mathcal{P}_i^*||\mathcal{P}_j)+\mathcal{D}_{kl}(\mathcal{P}_i||\mathcal{P}^*_j)\\
                \mathcal{L}_{\hat{y}_i\neq\hat{y}_j}=H_{\sigma}(\mathcal{D}_{kl}(\mathcal{P}_i^*||\mathcal{P}_j))+H_{\sigma}(\mathcal{D}_{kl}(\mathcal{P}_i||\mathcal{P}_j^*))\\
                H_{\sigma}(x)=\max(0,\sigma-x),
                \label{equ:bce}
            \end{gather}
        where $\sigma$ is a hyperparameter and $\mathcal{P}^*$ denotes that $\mathcal{P}$ is assumed to be a constant for asymmetry.  $\mathcal{D}_{kl}$ denotes KL divergence. 
        The probability distribution $\mathcal{P}$ will be pulled closer or farther depending on whether the labels of the sample pairs are the same. 
        In each iteration, if the annotator finds new relations, the parameters of the softmax classifier are reinitialized to deal with the new relations.  Alg.\ref{alg:1} shows an algorithm flow that is able to clearly summarize the proposed method.
        
            \begin{algorithm}[t]
                \caption{ASCORE}
                \label{alg:1}
                \KwIn{
                    A open dataset $\mathcal{D}=\{x_i\}_{i=1}^N$
                }
                \Repeat{\text{convergence}}{
                    Perform the encoding step (sec. \ref{sec:encoding}). Get relation representation $\{\bm{h}_i\}_{i=1}^N$ for instances in $\mathcal{D}$\;
                    \If{The conditions for stopping annotation are not met}{
                        Perform the labeling step (sec. \ref{sec:labeling}). Get the $s^{\text{th}}$ active labeled $\mathcal{D}^*=\mathcal{D}^*\cup\mathcal{D}_s$.\;
                    }
                    Perform the learning step (sec. \ref{sec:learning}). Estimate pseudo labels and reliability scores. Use the corresponding loss for representation learning.
                }
                \textbf{Return} Discovered relation set $\mathcal{R}$ and the cluster assignment $\hat{y}_i\in \mathcal{R}$ of each instance $x_i\in\mathcal{D}$.
            \end{algorithm}

\section{Experimental Setup}
    \subsection{Datasets}
        Experiments are conducted on two standard and one constructed dataset. Note that the compared baselines follow different settings. As will be described in sec. \ref{sec:baselines}, RSN and RoCORE leverage labeled data of predefined relations, while RW-HAC and SelfORE follow unsupervised setting. To fairly compare all methods in a uniform setting, the first half of the relations in each dataset are held out as predefined relations. Specifically, in TACRED, 21 relations are held out, while in FewRel and FewRel-LT, the number is 40.

        \noindent\textbf{TACRED} \citep{zhang-etal-2017-position}. TACRED is a large-scale manually annotated RE dataset, covering 41 relations. Following \citet{wu2019open,hu-etal-2020-selfore,zhao2021rocore}, the instances labeled as \texttt{no\_relation} are removed and the rest are used for training and evaluation.

        \noindent\textbf{FewRel}  \citep{han-etal-2018-fewrel}. FewRel is a manually annotated dataset that contains 80 types of relations, each of which has 700 instances.  
        However, in real-world OpenRE scenarios, unseen relations in unlabeled data usually follow a long-tailed distribution. To eliminate this inconsistency and accurately evaluate model performance in real-world scenarios, we construct a long-tail FewRel dataset as follows.
        
        \noindent\textbf{FewRel-LT}.   The FewRel-LongTail dataset. We number the 40 unseen relations in FewRel from 0 to 39, and calculate the number of samples according to $y=\frac{700}{0.5*id+1}$. The number of samples in the predefined subset remains unchanged.
        
        \begin{table*}[t]
            \centering
            \resizebox{\linewidth}{!}{
            \begin{tabular}{llc ccc ccc c ccc}
            \toprule
            \multirow{2}{*}{\textbf{Dataset}} & \multirow{2}{*}{\textbf{Method}} & \multirow{2}{*}{\textbf{Setting}} &  \multicolumn{3}{c}{$B^3$} & \multicolumn{3}{c}{V-measure} & \multirow{2}{*}{ARI} & \multicolumn{3}{c}{Classification}\\
            \cline{4-9}\cline{11-13}
            & & &Prec. & Rec. & $F_1$ & Hom. & Comp. & $F_1$ & & Prec. & Rec. & $F_1$ \\
            \midrule
            \multirow{7}{*}{\textbf{TACRED}} 
            &RW-HAC \begin{minipage}{0.02\textwidth}\includegraphics[width=3.2mm, height=3.2mm]{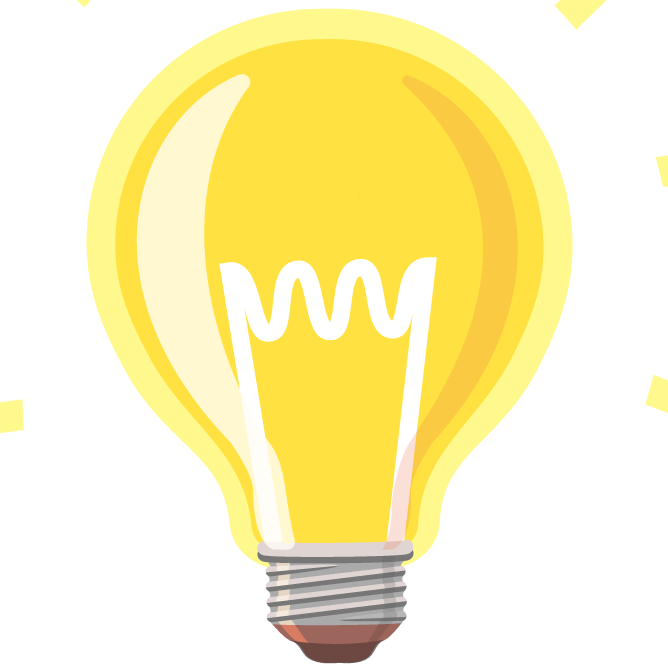}\end{minipage} \citep{elsahar2017unsupervised}& U &0.317 & 0.668 & 0.430 & 0.443 & 0.668 & 0.532 & 0.291 & 0.244 & 0.246 & 0.171  \\
            &SelfORE \begin{minipage}{0.02\textwidth}\includegraphics[width=3.2mm, height=3.2mm]{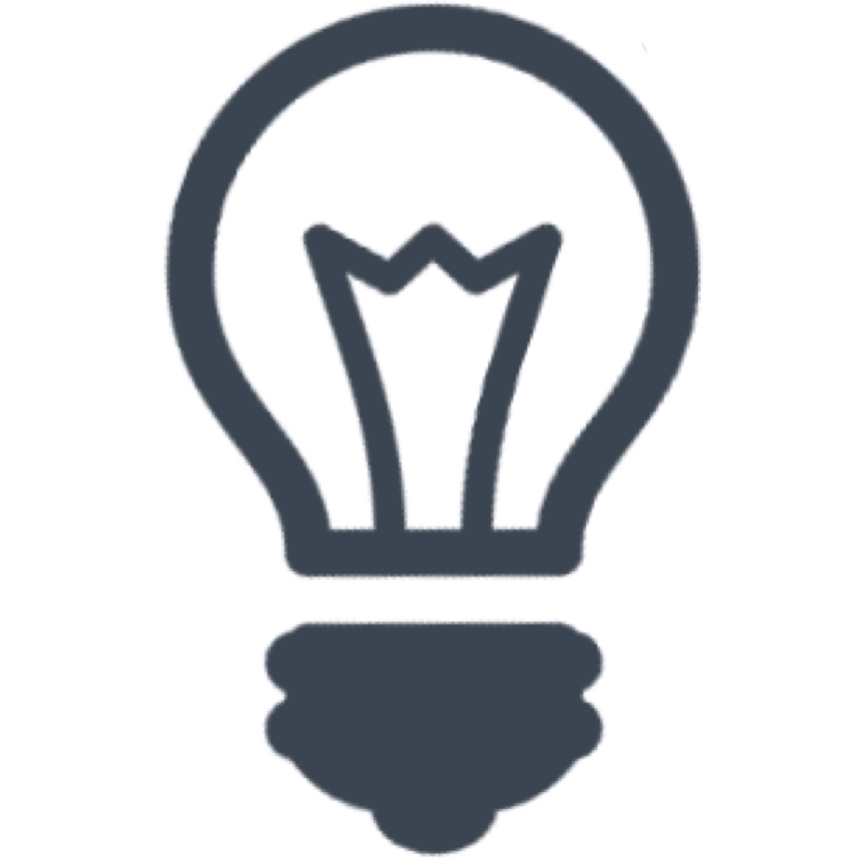}\end{minipage} \citep{hu-etal-2020-selfore} & U &0.339 & 0.759 & 0.469 & 0.468 & 0.809 & 0.593 & 0.412 &     0.121 & 0.244 & 0.160 \\
            &SelfORE \begin{minipage}{0.02\textwidth}\includegraphics[width=3.2mm, height=3.2mm]{open.png}\end{minipage} \citep{hu-etal-2020-selfore} & U &0.517 & 0.441 & 0.476 & 0.631 & 0.600 & 0.615 & 0.434 &  0.343 & 0.396 & 0.360 \\
            \cline{2-13}
            &RSN \begin{minipage}{0.02\textwidth}\includegraphics[width=3.2mm, height=3.2mm]{close.png}\end{minipage} \citep{wu2019open}& P &0.312 & 0.807 & 0.451 & 0.445 & 0.768 & 0.563 & 0.354 & 0.149 & 0.118 & 0.225\\
            &RoCORE\begin{minipage}{0.02\textwidth}\includegraphics[width=3.2mm, height=3.2mm]{close.png}\end{minipage} \citep{zhao2021rocore} & P &0.361 & 0.930 & 0.520 & 0.525 & 0.903 & 0.664 & 0.447 & 0.116 & 0.247 & 0.152\\
            &RoCORE\begin{minipage}{0.02\textwidth}\includegraphics[width=3.2mm, height=3.2mm]{open.png}\end{minipage} \citep{zhao2021rocore}& P &0.696 & 0.685 & 0.690 & 0.786 & 0.786 & 0.787 & 0.640 & 0.547 & 0.594 & 0.563 \\
            \rowcolor{gray!20}\cellcolor{white}&\textbf{Ours} \begin{minipage}{0.02\textwidth}\includegraphics[width=3.2mm, height=3.2mm]{close.png}\end{minipage} & A & 0.742 & 0.821 & \textbf{0.780} & 0.807 & 0.856 & \textbf{0.831} & \textbf{0.781} & 0.698 & 0.715 & \textbf{0.699}\\
            \hline \hline
            \multirow{7}{*}{\textbf{FewRel}}
            &RW-HAC \begin{minipage}{0.02\textwidth}\includegraphics[width=3.2mm, height=3.2mm]{open.png}\end{minipage} \citep{elsahar2017unsupervised}& U & 0.175 & 0.367 & 0.237 & 0.357 & 0.463 & 0.403 & 0.108 & 0.251 & 0.264 & 0.216\\
            &SelfORE \begin{minipage}{0.02\textwidth}\includegraphics[width=3.2mm, height=3.2mm]{close.png}\end{minipage} \citep{hu-etal-2020-selfore}& U &0.365  & 0.710 & 0.482 & 0.620 & 0.800 & 0.699 & 0.368 & 0.282 & 0.442 & 0.327\\
            &SelfORE \begin{minipage}{0.02\textwidth}\includegraphics[width=3.2mm, height=3.2mm]{open.png}\end{minipage} \citep{hu-etal-2020-selfore}& U &0.527  & 0.552 & 0.539 & 0.728 & 0.736 & 0.732 & 0.517 & 0.604 & 0.632 & 0.600\\
            \cline{2-13}
            &RSN \begin{minipage}{0.02\textwidth}\includegraphics[width=3.2mm, height=3.2mm]{close.png}\end{minipage} \citep{wu2019open}& P & 0.174 & 0.640 & 0.274 & 0.389 & 0.659 & 0.489 & 0.173 & 0.112 & 0.239 & 0.134 \\
            &RoCORE\begin{minipage}{0.02\textwidth}\includegraphics[width=3.2mm, height=3.2mm]{close.png}\end{minipage}\citep{zhao2021rocore}& P & 0.446 & 0.901 & 0.600 & 0.701  & 0.922 & 0.797 & 0.448 & 0.320 & 0.476 & 0.358\\
            &RoCORE\begin{minipage}{0.02\textwidth}\includegraphics[width=3.2mm, height=3.2mm]{open.png}\end{minipage}\citep{zhao2021rocore}& P & 0.806 & 0.843 & \textbf{0.824} & 0.883  & 0.896 & 0.889 & \textbf{0.807} & 0.827 & 0.868 & 0.837\\
            \rowcolor{gray!20}\cellcolor{white}&\textbf{Ours} \begin{minipage}{0.02\textwidth}\includegraphics[width=3.2mm, height=3.2mm]{close.png}\end{minipage} & A & 0.799 & 0.841 & 0.820 & 0.888 & 0.901 & \textbf{0.894} & 0.801 & 0.832 & 0.862 & \textbf{0.838}\\
            \hline \hline
            \multirow{7}{*}{\textbf{FewRel-LT}}
            &RW-HAC \begin{minipage}{0.02\textwidth}\includegraphics[width=3.2mm, height=3.2mm]{open.png}\end{minipage} \citep{elsahar2017unsupervised}& U & 0.255 & 0.322 & 0.285 & 0.379 & 0.421 & 0.399 & 0.145 & 0.190 & 0.176 & 0.160\\
            &SelfORE \begin{minipage}{0.02\textwidth}\includegraphics[width=3.2mm, height=3.2mm]{close.png}\end{minipage} \citep{hu-etal-2020-selfore}& U &0.266  & 0.633 & 0.374 & 0.466 & 0.676 & 0.552 & 0.290 & 0.079 & 0.154 & 0.099\\
            &SelfORE \begin{minipage}{0.02\textwidth}\includegraphics[width=3.2mm, height=3.2mm]{open.png}\end{minipage} \citep{hu-etal-2020-selfore}& U & 0.563 & 0.456 & 0.504 & 0.717 & 0.661 & 0.687 & 0.377 &  0.439 & 0.526 & 0.462 \\
            \cline{2-13}
            &RSN \begin{minipage}{0.02\textwidth}\includegraphics[width=3.2mm, height=3.2mm]{close.png}\end{minipage} \citep{wu2019open}& P & 0.211 & 0.500 & 0.297 & 0.350 & 0.510 & 0.415 & 0.193 & 0.098 & 0.173 & 0.117\\
            &RoCORE\begin{minipage}{0.02\textwidth}\includegraphics[width=3.2mm, height=3.2mm]{close.png}\end{minipage}\citep{zhao2021rocore}& P & 0.382 & 0.858 & 0.528 & 0.571 & 0.873 & 0.691 & 0.400 & 0.123 & 0.217 & 0.151\\
            &RoCORE\begin{minipage}{0.02\textwidth}\includegraphics[width=3.2mm, height=3.2mm]{open.png}\end{minipage}\citep{zhao2021rocore}& P & 0.662 & 0.717 & 0.689 & 0.800 & 0.801 & 0.800 & 0.581 & 0.507 & 0.538 & 0.517\\
            \rowcolor{gray!20}\cellcolor{white}&\textbf{Ours} \begin{minipage}{0.02\textwidth}\includegraphics[width=3.2mm, height=3.2mm]{close.png}\end{minipage} & A & 0.650 & 0.845 & \textbf{0.735} & 0.790 & 0.885 & \textbf{0.835} & \textbf{0.676} & 0.530 & 0.609 & \textbf{0.550}\\
            \bottomrule
            \end{tabular}
            }
            \caption{Main results on three relation extraction datasets. \begin{minipage}{0.02\textwidth}\includegraphics[width=3.2mm, height=3.2mm]{open.png}\end{minipage} and \begin{minipage}{0.02\textwidth}\includegraphics[width=3.2mm, height=3.2mm]{close.png}\end{minipage} represent that the number of relations is known and unknown, respectively (please look at appendix \ref{sec:app} for more details). U, P and A respectively indicate \textbf{U}nsupervised setting, supervised by \textbf{P}redefined relation setting and \textbf{A}ctively supervised setting. The proposed method outperforms the SOTA method and does not need to specify the number of clusters in advance.}
            \label{tab:main_res1}
        \end{table*}     
        
    \subsection{Compared Methods}
    \label{sec:baselines}
        To evaluate the effectiveness of the actively supervised clustering, the following SOTA two-stage OpenRE methods are used for comparison.
        
        \noindent\textbf{HAC with Re-weighted Word Embeddings (RW-HAC)} \citep{elsahar2017unsupervised}. A clustering-based OpenRE method. The model constructs relational feature based on entity types and the weighted sum of pretrained word embeddings.
        
        \noindent\textbf{Relational Siamese Network (RSN)} \citep{wu2019open}. This method learns similarity metrics of relations from labeled data of pre-defined relations and then transfers the relational knowledge to identify novel relations in unlabeled data. 
        
        
        \noindent\textbf{Self-supervised Feature Learning for OpenRE (SelfORE)} \citep{hu-etal-2020-selfore}. SelfORE exploits weak, self-supervised signals in pretrained language model for adaptive clustering on contextualized relational features. 
        
        \noindent\textbf{A Relation-oriented Clustering Method (RoCORE)} \citep{zhao2021rocore}. RoCORE leverages the labeled data of predefined relations to learn a clustering-friendly representation, which is used for new relations discovery.
        
        To show the superiority of the proposed labeling strategy, the actively supervised clustering is combined with the following classical active learning strategies for comparison. Specifically, \textbf{\textsc{Random}}, \textbf{\textsc{Confidence}} \citep{wang2014new}, \textbf{\textsc{Margin}} \citep{10.1007/11871842_40}, \textbf{\textsc{Entropy}} \citep{6889457} and \textbf{\textsc{Gradient}} \citep{DBLP:journals/corr/abs-1906-03671} are included. We provide a brief introduction to these methods in appendix \ref{sec:act}.
        
        
              
    \subsection{Implementation Details}
        Following \citet{hu-etal-2020-selfore} and \citet{wu2019open}, 20\% of the data in each dataset are held out for validation and hyperparameter selection. We use the Adam \citep{kingma2014adam} as the optimizer, with a learning rate of $1e-4$ and batch size of 100 for all datasets. The threshold $D_c$ is given by the value of an element ranked top 40\% in D from large to small. For each iteration, we label $B=20$ samples. $\xi$ is set to the value when the number of candidates is $1.2B$. Some important hyperparameters $r_h$ and $r_m$ are analyzed in sec. \ref{hyperparams}. For a fair comparison, all active strategies select the same number of key points for labeling. Specifically, 40, 80 and 80 key points are labeled on the three datasets TACRED, FewRel, and FewRel-LT respectively. All experiments are conducted with Pytorch 1.7.0, using an NVIDIA GeForce RTX 3090 with 24GB memory. 

\section{Main Results}
    \label{sec:main_res}
    Table \ref{tab:main_res1} shows the model performances on three datasets. In this section, the experiment focuses on the following two questions.
    \subsection{Does inaccurate estimation of the number of relations have an impact on clustering?}
            One drawback of the most existing two-stage OpenRE methods is that the number of clusters $K$ has to be given in advance, which is impractical in real applications. When $K$ is underestimated (from \begin{minipage}{0.02\textwidth}\includegraphics[width=3.2mm, height=3.2mm]{open.png}\end{minipage} to \begin{minipage}{0.02\textwidth}\includegraphics[width=3.2mm, height=3.2mm]{close.png}\end{minipage}), the clustering performance of the SOTA unsupervised method, SelfORE, on the three datasets decreases by an average of 7.13\%, while the same metric regarding RoCORE, the SOTA supervised method, is 18.10\%. Furthermore, it is observed an extremely unbalanced precision-recall value in the $B^3$ metric (much lower precision and higher recall), which indicates that the model tends to mix multiple relations in the same cluster. Obviously, such clustering results will have a negative impact on the relation labeling. In other words, it is difficult to determine which relation this cluster corresponds to. When $K$ is overestimated (due to space limitation, please look at table \ref{tab:main_res1_full} for results of overestimation), the same relation tends to be clustered into multiple subclusters. Repeated labeling of these subclusters brings a significant increase in human effort. In contrast, the ASCORE dynamically discovers relational clusters through active iteration, breaking the impractical assumption that $K$ is known in advance.
            
        \subsection{Is the actively supervised setting better than the two-stage setting?}
            The two settings are compared, from the following two perspectives.
            
            In terms of clustering performance, the actively labeled data can provide valuable supervision signals for clustering learning. Compared with RoCORE, a strong baseline supervised by predefined relations, the proposed method improves the four metrics by an average of
            $10.3\%$ and $5.2\%$ on long-tail TACRED and FewRel-LT, respectively. Long-tail relation distribution is very common in real world. On the uniform FewRel dataset, the ASCORE achieves comparable results. It is worth noting that the number of clusters $K$ has to be supplied for RoCORE. When $K$ is unknown, the improvement will be even enlarged.
            
            Regarding labeling costs, both settings are comparable. Note that in the main results, only two instances for each relation are labeled on average. For the two-stage methods, in order to label the relational semantics of a cluster, the annotator has to observe at least one sample. Obviously, the ASCORE does not lead to a significant increase in human efforts.

        \begin{table}
            \centering
            \resizebox{\linewidth}{!}{
            \begin{tabular}{lc ccc cc}
            \toprule
            \textbf{Dataset} & \textsc{Rand.} & \textsc{Conf.} & \textsc{Marg.} & \textsc{Entro.} & \textsc{Grad.} & \textbf{\textsc{Ours}}\\
            \midrule
            \textbf{TACRED} & 15 & 13 & 18 & 16 & 17 & 18\\
            \textbf{FewRel} & 40 & 34 & 40 & 37 & 40 & 40\\
            \textbf{FewRel-LT} & 27 & 19 & 25 & 26 & 30 & 35\\
            \textbf{ALL} & 82($\downarrow5.7\%$) & 66($\downarrow24.1\%$) & 83($\downarrow4.6\%$) & 79($\downarrow9.2\%$) & 87($0\%$) & \textbf{92}\textcolor{red}{($\uparrow5.8\%$)}\\
            \bottomrule
            \end{tabular}
            }
            \caption{\label{tab:diverse}
            The number of relations discovered by the labeling strategies. (\textsc{Grad.} as a reference point.)
            }
        \end{table} 
        \section{Analysis and Discussions}
        \subsection{Analysis on Labeling Strategy}
        Main results (sec. \ref{sec:main_res}) have shown the advantages of the proposed actively supervised clustering setting over the two-stage setting. 
        \textit{However, some serious readers may still think the comparison across settings is unfair}. To further reduce readers' concerns and show the effectiveness of our labeling strategy, we combine the actively supervised clustering settings with the various active labeling strategies and compare them in terms of relation discovery and clustering performance. Note that the number of key points selected by each strategy is the same (two points per relation on average). The results are shown in tab.\ref{tab:diverse} and tab. \ref{tab:main_res2}. It can be seen from tab. \ref{tab:diverse} that the proposed labeling strategy finds the most relations. Different from the classical strategies that focus only on improving the recognition of relations that have been discovered, the proposed strategy appropriately explores new relations by distance regularization, which is particularly beneficial for long-tail relation discovery in real applications. Additionally, tab. \ref{tab:main_res2} shows that this strategy is also the best in terms of clustering performance. Benefitting from assigning reasonable loss functions to pseudo labels with different reliability, more pseudo labels can be used for learning without significantly increasing the risk of over-fitting noise.
  
        \begin{table}
            \centering
            \resizebox{\linewidth}{!}{
            \begin{tabular}{ll cccc}
            \toprule
            \textbf{Dataset} & \textbf{Method} & $B^3$ & V-measure & ARI & Classification\\
            \midrule
            \multirow{6}{*}{\textbf{TACRED}} 
            &\textsc{Random} & 0.737 & 0.800 & 0.662 & 0.464\\
            &\textsc{Confidence} & 0.671 & 0.752 & 0.598 & 0.408\\
            &\textsc{Margin} & 0.709 & 0.787 & 0.628 & 0.524\\
            &\textsc{Entropy} & 0.702 & 0.790 & 0.633 & 0.502\\
            &\textsc{Gradient} & 0.767& 0.831 & 0.725 & 0.670\\
            \rowcolor{gray!20}\cellcolor{white}&\textbf{Ours} & \textbf{0.780} & \textbf{0.851} & \textbf{0.781} & \textbf{0.699}\\
            \hline
            \multirow{6}{*}{\textbf{FewRel}}
            &\textsc{Random} & 0.808 & 0.882 & 0.787 & 0.813\\
            &\textsc{Confidence} & 0.752 & 0.851 & 0.660 & 0.701\\
            &\textsc{Margin} & 0.817 & 0.881 & 0.796 & 0.831\\
            &\textsc{Entropy} & 0.781 & 0.819 & 0.779 & 0.743\\
            &\textsc{Gradient} & 0.814 & 0.884 & 0.790 & 0.827\\
            \rowcolor{gray!20}\cellcolor{white}&\textbf{Ours} & \textbf{0.820} & \textbf{0.894} & \textbf{0.801} & \textbf{0.838}\\
            \hline
            \multirow{6}{*}{\textbf{FewRel-LT}}
            &\textsc{Random} & 0.716 & 0.814 & 0.670 & 0.486\\
            &\textsc{Confidence} & 0.346 & 0.514 & 0.217 & 0.336\\
            &\textsc{Margin} & 0.696 & 0.806 & 0.647 & 0.481\\
            &\textsc{Entropy} & 0.721 & 0.824 & 0.664 & 0.481\\
            &\textsc{Gradient} & 0.719 & 0.797 & 0.649 & 0.498\\
            \rowcolor{gray!20}\cellcolor{white}&\textbf{Ours} & \textbf{0.735} & \textbf{0.835} & \textbf{0.676} & \textbf{0.550}\\
            \bottomrule
            \end{tabular}
            }
            \caption{Comparison results of labeling strategies on three datasets.} 
            \label{tab:main_res2}
        \end{table}   
    \subsection{The Effect of Numbers of Actively Labeled Instances}
\begin{figure}[t]
            \includegraphics[width=\columnwidth]{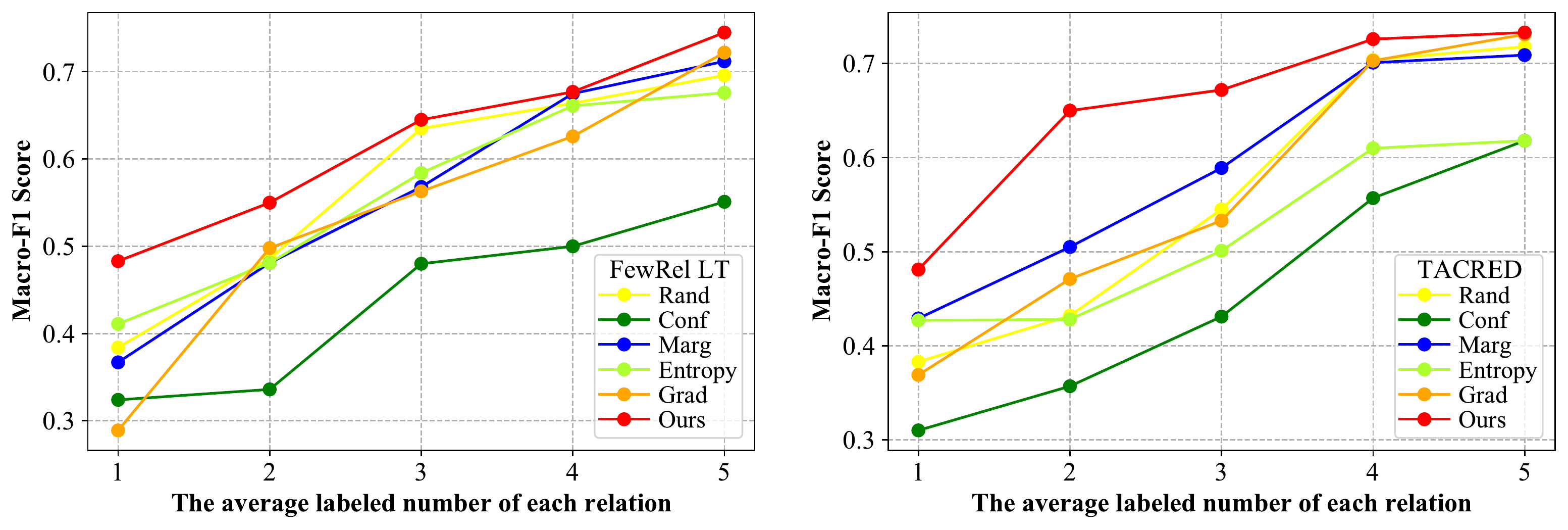}
            \caption{Performance with different active labeled instances.}
            \label{fig:num_query}
        \end{figure}

        In order to more comprehensively evaluate the labeling strategies, experiments are conducted to compare these labeling strategies by varying the number of actively labeled instances, $N^*$. Figure \ref{fig:num_query} shows the effect of $N^*$. Surprisingly, it is found that the random strategy is a very competitive baseline that beats most of the classical labeling strategies given different $N^*$. This suggests that classical labeling strategies may be better at tasks with known and fixed categories. Although the proposed strategy consistently outperforms all baselines, it is obvious that it has not been fully optimized. It is the authors' belief that it is sufficient to serve as a reasonable baseline for the actively supervised clustering setting and proceeding some useful guidance for future research in this field. Additionally, with the increase of $N^*$, the performance of the model is improved, but the growth rate is gradually slow. It means that the cost performance of human effort gradually decreases. Therefore, for users with limited budgets, it is also a good choice to discover the primary relations through only a few queries.
\subsection{Hyperparameter Analysis}
        \label{hyperparams}
        \begin{figure}[t]
            \includegraphics[width=\columnwidth]{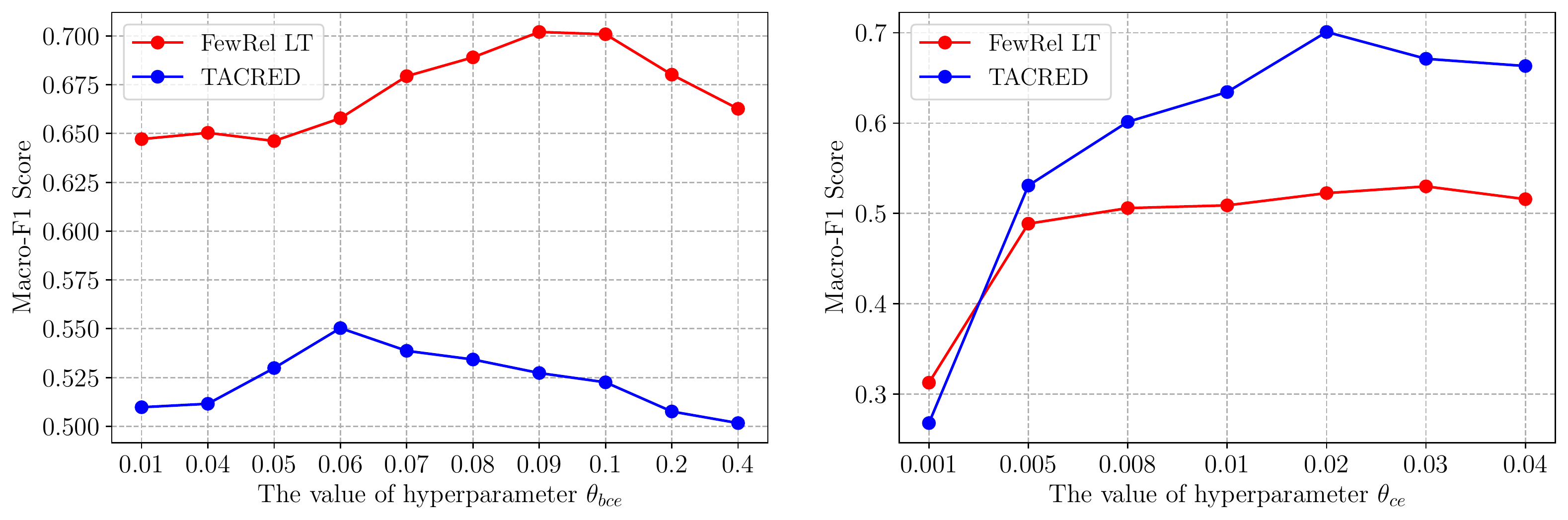}
            \caption{Performance with different hyperparameter settings.}
            \label{fig:hyper}
        \end{figure}
        In this section, we study the effects of reliability threshold $r_h$ and $r_m$ on optimization. Their values are given by the values of the elements ranked $\theta_{ce}\%$ and $\theta_{bce}\%$ from small to large. From Figure \ref{fig:hyper} it is possible see that: (1) When $\theta_{ce}$ and $\theta_{bce}$ gradually increase from a small value, more training data are used for model optimization, and the performance of the model gradually improves. (2) When the value exceeds a certain threshold, further increasing the $\theta_{ce}$ and $\theta_{bce}$ will introduce more errors into the optimization, which degrades the performance of the model. (3) Compared with $\mathcal{L}_{bce}$, $\mathcal{L}_{ce}$ loss will make the model converge faster, so its optimal threshold of $\mathcal{L}_{ce}$ should be less than that of $\mathcal{L}_{bce}$ to prevent the overfitting of wrong labels.
\section{Conclusions}

In this work, we present a new setting, named actively supervised clustering for OpenRE, which provides the necessary guidance for clustering without a significant increase in human efforts. Along with this setting, a labeling strategy tailored for clustering is proposed, maximizing the clustering performance while discovering as many relations as possible. Different loss functions are assigned to pseudo labels with different reliability, which mitigate the risk of over-fitting to noise in pseudo labels. Experimental results show that this method significantly outperforms the existing two-stage methods for OpenRE.
 
\section*{Limitations}
Considering that the golden labels of all instances have been given in the datasets, we directly use these labels as manual labels without performing the manual labeling process. The practice implicitly assumes that all the manual labels are correct. However, with the increase of labeling scale, problems such as (1) inconsistent labeling granularity across annotators and (2) noise in manual labels gradually emerge. How to effectively improve the labeling quality and the robustness of the clustering model are worthy of attention.

\section*{Acknowledgements}
The authors wish to thank the anonymous reviewers for their helpful comments. This work was partially funded by National Natural Science Foundation of China (No.62076069,62206057,61976056), Shanghai Rising-Star Program (23QA1400200), and Natural Science Foundation of Shanghai (23ZR1403500).
\bibliography{anthology,custom}
\bibliographystyle{acl_natbib}

\appendix

\section{Appendix}
\begin{table*}
            \centering
            \resizebox{\linewidth}{!}{
            \begin{tabular}{llc ccc ccc c ccc}
            \toprule
            \multirow{2}{*}{\textbf{Dataset}} & \multirow{2}{*}{\textbf{Method}} & \multirow{2}{*}{\textbf{Setting}} &  \multicolumn{3}{c}{$B^3$} & \multicolumn{3}{c}{V-measure} & \multirow{2}{*}{ARI} & \multicolumn{3}{c}{Classification}\\
            \cline{4-9}\cline{11-13}
            & & &Prec. & Rec. & $F_1$ & Hom. & Comp. & $F_1$ & & Prec. & Rec. & $F_1$ \\
            \midrule
            \multirow{9}{*}{\textbf{TACRED}} 
            &RW-HAC \begin{minipage}{0.02\textwidth}\includegraphics[width=3.2mm, height=3.2mm]{open.png}\end{minipage} \citep{elsahar2017unsupervised}& U &0.317 & 0.668 & 0.430 & 0.443 & 0.668 & 0.532 & 0.291 & 0.244 & 0.246 & 0.171  \\
            &SelfORE \begin{minipage}{0.02\textwidth}\includegraphics[width=3.2mm, height=3.2mm]{close.png}\end{minipage}- \citep{hu-etal-2020-selfore} & U &0.339 & 0.759 & 0.469 & 0.468 & 0.809 & 0.593 & 0.412 &    0.121 & 0.244 & 0.160 \\
            &SelfORE \begin{minipage}{0.02\textwidth}\includegraphics[width=3.2mm, height=3.2mm]{close.png}\end{minipage}+ \citep{hu-etal-2020-selfore} & U &0.575 & 0.251 & 0.349 & 0.680 & 0.522 & 0.591 & 0.290 &    0.469 & 0.481 & 0.231 \\
            &SelfORE \begin{minipage}{0.02\textwidth}\includegraphics[width=3.2mm, height=3.2mm]{open.png}\end{minipage} \citep{hu-etal-2020-selfore} & U &0.517 & 0.441 & 0.476 & 0.631 & 0.600 & 0.615 & 0.434 &  0.343 & 0.396 & 0.360 \\
            \cline{2-13}
            &RSN \begin{minipage}{0.02\textwidth}\includegraphics[width=3.2mm, height=3.2mm]{close.png}\end{minipage} \citep{wu2019open}& P &0.312 & 0.807 & 0.451 & 0.445 & 0.768 & 0.563 & 0.354 & 0.149 & 0.118 & 0.225\\
            &RoCORE\begin{minipage}{0.02\textwidth}\includegraphics[width=3.2mm, height=3.2mm]{close.png}\end{minipage}- \citep{zhao2021rocore} & P &0.361 & 0.930 & 0.520 & 0.525 & 0.903 & 0.664 & 0.447 & 0.116 & 0.247 & 0.152\\
            &RoCORE\begin{minipage}{0.02\textwidth}\includegraphics[width=3.2mm, height=3.2mm]{close.png}\end{minipage}+ \citep{zhao2021rocore} & P &0.714 & 0.530 & 0.608 & 0.805 & 0.698 & 0.748 & 0.552 & 0.612 & 0.649 & 0.311\\
            &RoCORE\begin{minipage}{0.02\textwidth}\includegraphics[width=3.2mm, height=3.2mm]{open.png}\end{minipage} \citep{zhao2021rocore}& P &0.696 & 0.685 & 0.690 & 0.786 & 0.786 & 0.787 & 0.640 & 0.547 & 0.594 & 0.563 \\
            \rowcolor{gray!20}\cellcolor{white}&\textbf{Ours} \begin{minipage}{0.02\textwidth}\includegraphics[width=3.2mm, height=3.2mm]{close.png}\end{minipage} & A & 0.742 & 0.821 & \textbf{0.780} & 0.807 & 0.856 & \textbf{0.831} & \textbf{0.781} & 0.698 & 0.715 & \textbf{0.699}\\
            \hline \hline
            \multirow{9}{*}{\textbf{FewRel}}
            &RW-HAC \begin{minipage}{0.02\textwidth}\includegraphics[width=3.2mm, height=3.2mm]{open.png}\end{minipage} \citep{elsahar2017unsupervised}& U & 0.175 & 0.367 & 0.237 & 0.357 & 0.463 & 0.403 & 0.108 & 0.251 & 0.264 & 0.216\\
            &SelfORE \begin{minipage}{0.02\textwidth}\includegraphics[width=3.2mm, height=3.2mm]{close.png}\end{minipage}- \citep{hu-etal-2020-selfore}& U &0.365  & 0.710 & 0.482 & 0.620 & 0.800 & 0.699 & 0.368 & 0.282 & 0.442 & 0.327\\
            &SelfORE \begin{minipage}{0.02\textwidth}\includegraphics[width=3.2mm, height=3.2mm]{close.png}\end{minipage}+ \citep{hu-etal-2020-selfore}& U &0.639  & 0.400 & 0.492 & 0.793 & 0.681 & 0.733 & 0.492 & 0.733 & 0.744 & 0.365\\
            &SelfORE \begin{minipage}{0.02\textwidth}\includegraphics[width=3.2mm, height=3.2mm]{open.png}\end{minipage} \citep{hu-etal-2020-selfore}& U &0.527  & 0.552 & 0.539 & 0.728 & 0.736 & 0.732 & 0.517 & 0.604 & 0.632 & 0.600\\
            \cline{2-13}
            &RSN \begin{minipage}{0.02\textwidth}\includegraphics[width=3.2mm, height=3.2mm]{close.png}\end{minipage} \citep{wu2019open}& P & 0.174 & 0.640 & 0.274 & 0.389 & 0.659 & 0.489 & 0.173 & 0.112 & 0.239 & 0.134 \\
            &RoCORE\begin{minipage}{0.02\textwidth}\includegraphics[width=3.2mm, height=3.2mm]{close.png}\end{minipage}-\citep{zhao2021rocore}& P & 0.446 & 0.901 & 0.600 & 0.701  & 0.922 & 0.797 & 0.448 & 0.320 & 0.476 & 0.358\\
            &RoCORE\begin{minipage}{0.02\textwidth}\includegraphics[width=3.2mm, height=3.2mm]{close.png}\end{minipage}+\citep{zhao2021rocore}& P & 0.824 & 0.656 & 0.730 & 0.896  & 0.814 & 0.853 & 0.739 & 0.882 & 0.881 & 0.439\\
            &RoCORE\begin{minipage}{0.02\textwidth}\includegraphics[width=3.2mm, height=3.2mm]{open.png}\end{minipage}\citep{zhao2021rocore}& P & 0.806 & 0.843 & \textbf{0.824} & 0.883  & 0.896 & 0.889 & \textbf{0.807} & 0.827 & 0.868 & 0.837\\
            \rowcolor{gray!20}\cellcolor{white}&\textbf{Ours} \begin{minipage}{0.02\textwidth}\includegraphics[width=3.2mm, height=3.2mm]{close.png}\end{minipage} & A & 0.799 & 0.841 & 0.820 & 0.888 & 0.901 & \textbf{0.894} & 0.801 & 0.832 & 0.862 & \textbf{0.838}\\
            \hline \hline
            \multirow{9}{*}{\textbf{FewRel-LT}}
            &RW-HAC \begin{minipage}{0.02\textwidth}\includegraphics[width=3.2mm, height=3.2mm]{open.png}\end{minipage} \citep{elsahar2017unsupervised}& U & 0.255 & 0.322 & 0.285 & 0.379 & 0.421 & 0.399 & 0.145 & 0.190 & 0.176 & 0.160\\
            &SelfORE \begin{minipage}{0.02\textwidth}\includegraphics[width=3.2mm, height=3.2mm]{close.png}\end{minipage}- \citep{hu-etal-2020-selfore}& U &0.266  & 0.633 & 0.374 & 0.466 & 0.676 & 0.552 & 0.290 & 0.079 & 0.154 & 0.099\\
            &SelfORE \begin{minipage}{0.02\textwidth}\includegraphics[width=3.2mm, height=3.2mm]{close.png}\end{minipage}+ \citep{hu-etal-2020-selfore}& U &0.641  & 0.298 & 0.407 & 0.771 & 0.589 & 0.668 & 0.370 & 0.589 & 0.647 & 0.305\\
            &SelfORE \begin{minipage}{0.02\textwidth}\includegraphics[width=3.2mm, height=3.2mm]{open.png}\end{minipage} \citep{hu-etal-2020-selfore}& U & 0.563 & 0.456 & 0.504 & 0.717 & 0.661 & 0.687 & 0.377 &  0.439 & 0.526 & 0.462 \\
            \cline{2-13}
            &RSN \begin{minipage}{0.02\textwidth}\includegraphics[width=3.2mm, height=3.2mm]{close.png}\end{minipage} \citep{wu2019open}& P & 0.211 & 0.500 & 0.297 & 0.350 & 0.510 & 0.415 & 0.193 & 0.098 & 0.173 & 0.117\\
            &RoCORE\begin{minipage}{0.02\textwidth}\includegraphics[width=3.2mm, height=3.2mm]{close.png}\end{minipage}-\citep{zhao2021rocore}& P & 0.382 & 0.858 & 0.528 & 0.571 & 0.873 & 0.691 & 0.400 & 0.123 & 0.217 & 0.151\\
            &RoCORE\begin{minipage}{0.02\textwidth}\includegraphics[width=3.2mm, height=3.2mm]{close.png}\end{minipage}+\citep{zhao2021rocore}& P & 0.714 & 0.530 & 0.608 & 0.805 & 0.698 & 0.748 & 0.552 & 0.612 & 0.649 & 0.311\\
            &RoCORE\begin{minipage}{0.02\textwidth}\includegraphics[width=3.2mm, height=3.2mm]{open.png}\end{minipage}\citep{zhao2021rocore}& P & 0.662 & 0.717 & 0.689 & 0.800 & 0.801 & 0.800 & 0.581 & 0.507 & 0.538 & 0.517\\
            \rowcolor{gray!20}\cellcolor{white}&\textbf{Ours} \begin{minipage}{0.02\textwidth}\includegraphics[width=3.2mm, height=3.2mm]{close.png}\end{minipage} & A & 0.650 & 0.845 & \textbf{0.735} & 0.790 & 0.885 & \textbf{0.835} & \textbf{0.676} & 0.530 & 0.609 & \textbf{0.550}\\
            \bottomrule
            \end{tabular}
            }
            \caption{Main results on three relation extraction datasets. \begin{minipage}{0.02\textwidth}\includegraphics[width=3.2mm, height=3.2mm]{open.png}\end{minipage} and \begin{minipage}{0.02\textwidth}\includegraphics[width=3.2mm, height=3.2mm]{close.png}\end{minipage} represent that the number of relation types in unlabeled data is correctly and incorrectly estimated, respectively. In addition, - and + denotes the underestimation and overestimation, respectively.} 
            \label{tab:main_res1_full}
        \end{table*}    
\subsection{Compared Active Labeling Strategy}
\label{sec:act}
To show the superiority of the proposed labeling strategy, the actively supervised clustering is combined with the following classical active learning strategies for comparison.
        \noindent\textbf{\textsc{Random}} The naive baseline of randomly selecting $k$ samples to query labels.
        
        \noindent\textbf{\textsc{Confidence}} \citep{wang2014new} An uncertainty-based active learning algorithm that select $k$ samples with smallest predicted class probability $\max\{f_\theta(x)_i\}_{i=1,...,C}$
        
        \noindent\textbf{\textsc{Margin}} \citep{10.1007/11871842_40} An uncertainty-based active learning algorithm that selects the bottom $k$ sorted according to the example's multiclass margin, defined as $f_\theta(x)_{\hat{y}}-f_\theta(x)_{y^\prime}$, where $\hat{y}$ and $y^\prime$ are the indices of the largest and second largest entries of $f_\theta(x)$. 
        
        \noindent\textbf{\textsc{Entropy}}\citep{6889457} An uncertainty-based active learning algorithm that selects the top $k$ samples according to the entropy of the sample's class distribution.
        
        \noindent\textbf{\textsc{Gradient}}\citep{DBLP:journals/corr/abs-1906-03671} A loss based active learning algorithm. The uncertainty is measured as the gradient magnitude with respect to parameters in the output layer.

\subsection{Additional Results}
\label{sec:app}
    In this section, more detailed experimental settings and experimental results are given. Specifically, in TACRED dataset, 21 relations are held out as open relations to be discovered. In the underestimation (resp. overestimation) setting, we assume that the number of clusters is 10 (resp. 40). In FewRel and FewRel-LT datasets, 40 relations are held out as open relations and we assume that the number of clusters is 20 (resp. 80) for underestimation (resp. overestimation) setting. The results of the two settings are listed in tab. \ref{tab:main_res1_full}. When $K$ is underestimated, it is observed that the precision of $B^3$ is far lower than recall, which indicates that the model tends to mix multiple relations in the same cluster. When $K$ is overestimated, the recall is far lower than precision, which indicates that the same relation tends to be clustered into multiple subclusters. Although the $F_1$ of $B^3$ metric seems to be tolerable, such imbalance clustering assignments cause great difficulties in relation labeling. If a cluster contains more than one relation, labeling the cluster as any relation will lead to the misidentification of other relations. If a relation is clustered into multiple sub-clusters, the annotators have to label the same relation repeatedly, which leads to a significant increase in labeling costs.

\end{document}